\pdfoutput=1

\documentclass[11pt]{article}

\usepackage{ACL2023}

\usepackage{times}
\usepackage{latexsym}

\usepackage[T1]{fontenc}

\usepackage[utf8]{inputenc}

\usepackage{microtype}

\usepackage{inconsolata}

\title{\ts{NewsDialogues}: Towards Proactive News Grounded Conversation}

\author{
Siheng Li$^{1}$\thanks{$^*$ This work is done when Siheng Li is an intern at Huawei Noah's Ark Lab.},
Yichun Yin$^{2}$, 
Cheng Yang$^{1}$,
Wangjie Jiang$^{1}$,
Yiwei Li$^{3}$
\\
{\bf Zesen Cheng}$^{4}$, 
{\bf Lifeng Shang}$^{2}$, 
{\bf Xin Jiang}$^{2}$, 
{\bf Qun Liu}$^{2}$, 
{\bf Yujiu Yang}$^{1}$\thanks{$^\dagger$ Corresponding author.} \\ 
$^1$Shenzhen International Graduate School, Tsinghua University \\
$^2$Huawei Noah’s Ark Lab, $^3$Beijing Institute of Technology, $^4$Peking University \\
lisiheng21@mails.tsinghua.edu.cn \\
\{yinyichun, shang.lifeng, jiang.xin, qun.liu\}@huawei.com \\
yang.yujiu@sz.tsinghua.edu.cn
}

\usepackage{array}
\usepackage{pifont}
\usepackage{tabularx}
\usepackage{adjustbox}
\usepackage{multirow}
\usepackage{enumitem}
\usepackage{xspace}
\usepackage{tcolorbox}
\usepackage{booktabs,amsfonts,dcolumn}
\usepackage{hyperref}
\usepackage{url}
\usepackage{amsmath,amsthm,amsfonts,amssymb,bm,stmaryrd,bbm}
\usepackage[noorphans,vskip=0.75ex,leftmargin=2ex]{quoting}
\usepackage{graphicx}
\usepackage{CJKutf8}
\usepackage{enumerate}
\usepackage{hyperref}

\newcommand\ti[1]{\textit{#1}}
\newcommand\ts[1]{\textsc{#1}}
\newcommand\tb[1]{\textbf{#1}}
\newcommand\tc[2]{\textcolor{#1}{#2}}

\newcommand\ntb[1]{\noindent \textbf{#1}\quad}

\newcommand{\cls}{\texttt{[CLS]}}
\newcommand{\vsd}{\vspace{-5pt}}

\definecolor{hong}{HTML}{F6416C}
\definecolor{lan}{HTML}{4D77FF}
\definecolor{lv}{HTML}{00B8A9}
\definecolor{huang}{HTML}{FF8E00}

\newcommand\cmark {\textcolor{green}{\ding{52}}}
\newcommand\xmark {\textcolor{red}{\ding{55}}}

\begin{document}
\maketitle

\begin{abstract}

Hot news is one of the most popular topics in daily conversations.
However, news grounded conversation has long been stymied by the lack of well-designed task definition and scarce data. 
In this paper, we propose a novel task, Proactive News Grounded Conversation, in which a dialogue system can proactively lead the conversation based on some key topics of the news.
In addition, both information-seeking and chit-chat scenarios are included realistically, where the user may ask a series of questions about the news details or express their opinions and be eager to chat.
To further develop this novel task, we collect a human-to-human Chinese dialogue dataset \ts{NewsDialogues}, which includes 1K conversations with a total of 14.6K utterances and detailed annotations for target topics and knowledge spans.
Furthermore, we propose a method named Predict-Generate-Rank, consisting of a 
generator for grounded knowledge prediction and response generation, and a ranker for the ranking of multiple responses to alleviate the exposure bias.
We conduct comprehensive experiments to demonstrate the effectiveness of the proposed method and further present several key findings and challenges to prompt future research.\footnote{\ The project repository is available at \url{https://github.com/SihengLi99/NewsDialogues}.}

\end{abstract}
\section{Introduction}
News, especially hot news, is widely discussed in daily conversations, enabling people to connect to others and engage with the public issues they encounter in everyday life \cite{swart2017repositioning}. 
However, due to the lack of well-designed task definition and the scarcity of training data,
news grounded conversation has almost been neglected in dialogue system research \cite{huang2020challenges, ni2021recent, thoppilan2022lamda}. 

\begin{table}[t]
    \begin{center}
    \centering
    \small
    \resizebox{\linewidth}{!}{%
    \begin{tabular}{lcccc}
    \toprule
       \tb{Dataset} & \tb{Domain} & \tb{A-p} & \tb{C-c} & \tb{I-s} \\
    \midrule
    \midrule
        CMU DoG \citep{DBLP:conf/emnlp/ZhouPB18} & Film & \xmark & \cmark & \cmark \\
        India Dog \citep{DBLP:conf/emnlp/MogheABK18} & Film & \xmark & \cmark & \xmark \\
        QuAC \citep{DBLP:conf/emnlp/ChoiHIYYCLZ18} & Wikipedia & \xmark & \xmark & \cmark \\ 
        CoQA \citep{DBLP:journals/tacl/ReddyCM19} & Multi-Domain & \xmark & \xmark & \cmark \\
        WoW \citep{DBLP:conf/iclr/DinanRSFAW19} & Wikipedia & \xmark & \cmark & \xmark \\
        doc2dial \citep{DBLP:conf/emnlp/FengWGPJL20} & Service & \xmark & \xmark & \cmark \\
        WikiDialog \citep{DBLP:conf/icml/DaiCZARGG22} & Wikipedia & \xmark & \xmark & \cmark \\
        {\textsc{InSCIt}\xspace} \citep{DBLP:journals/corr/abs-2207-00746} & Wikipedia & \xmark & \xmark & \cmark \\
        \ts{NewsDialogues} (Ours) & News & \cmark & \cmark & \cmark \\
    \bottomrule
    \end{tabular}
    }
    \end{center}

    \caption{
        The differences between \ts{NewsDialogues} and other document-grounded dialogue datasets.
        A-p represents the modeling of agent proactivity,
        C-c and I-s denotes whether the conversations focus on chit-chat scenario and information-seeking scenario respectively.
    }
    \label{tab:dataset_comparison}
    \vspace{-5pt}
\end{table}

\begin{figure*}
    \centering
    \includegraphics[width=\textwidth]{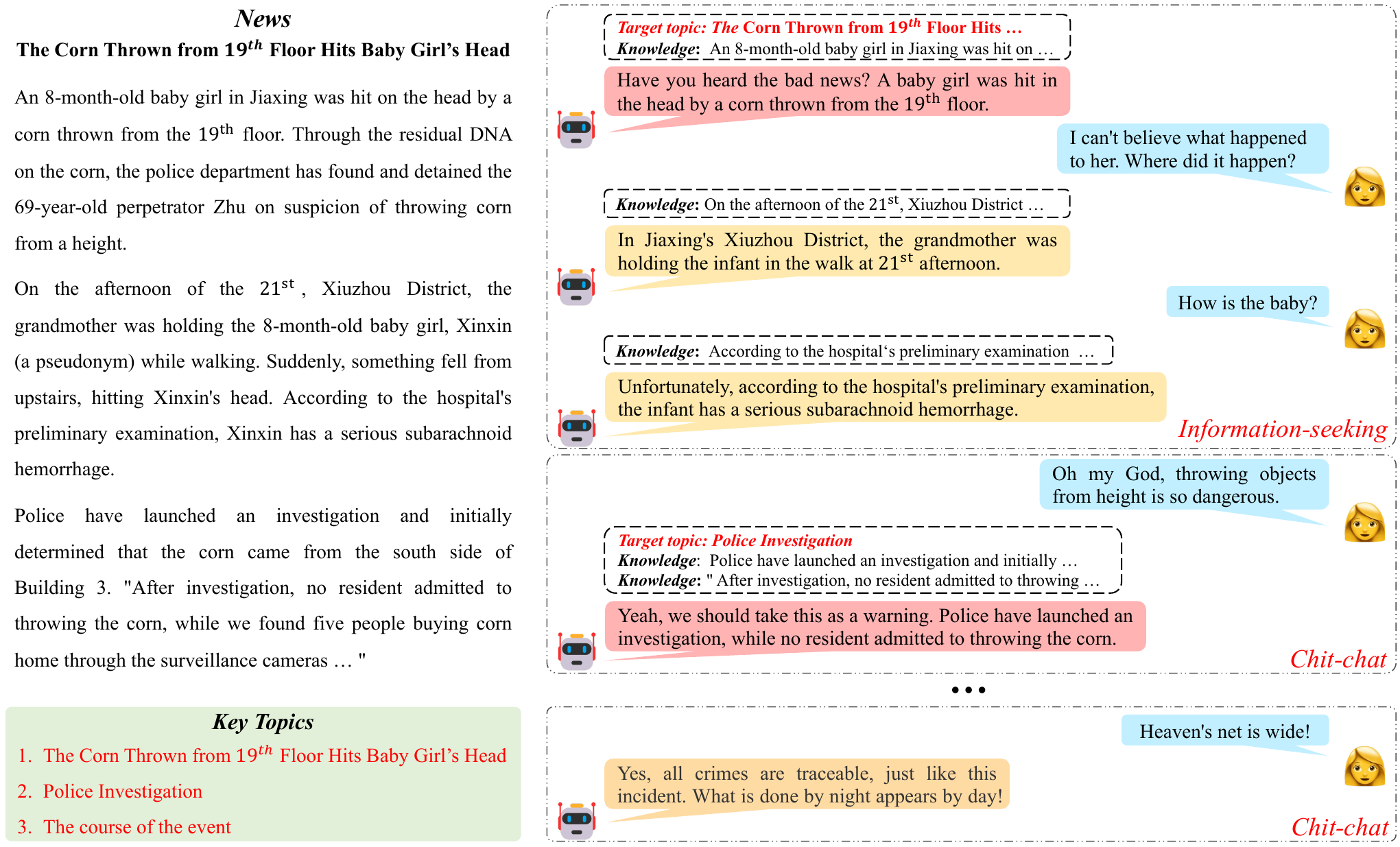}
    \caption{An example of \ts{NewsDialogues}. We translate the original Chinese dialogue to English version for reading convenience. Notice that some content is omitted as the original version is too long, please refer to the original example in Appendix Figure \ref{fig:chinese_example}.}
    \label{fig:example}
\end{figure*}

To pursue news grounded conversation, a natural idea is to refer to existing document-grounded conversations.
However, there are two major differences.
First, as news articles can be long, complex, and time-consuming for human reading, it is important for the dialogue system to be proactive, which means that it can actively introduce news content during the conversation.
Therefore, users know more about the news, and the conversations are more interactive and in-depth.
However, traditional document-grounded dialogue datasets rarely consider this proactivity explicitly, and the conversations are more user-driven.
For example, in \ti{QuAC} \cite{DBLP:conf/emnlp/ChoiHIYYCLZ18}, \ti{doc2dial} \citep{DBLP:conf/emnlp/FengWGPJL20}, and \ti{WikiDialog} \cite{DBLP:conf/icml/DaiCZARGG22}, systems mostly respond to user questions passively based on the documents.
Second, both chit-chat and information-seeking scenarios \citep{stede2004information, DBLP:conf/emnlp/ChoiHIYYCLZ18} are indispensable for news grounded conversation.
Users may ask a series of questions about the news details curiously or express their opinions and be eager to chat.
However, existing document-grounded conversations mostly focus on a single scenario of chit-chat or information-seeking rather than both.
The work of \citet{DBLP:conf/emnlp/ChoiHIYYCLZ18, DBLP:conf/emnlp/FengWGPJL20, DBLP:conf/icml/DaiCZARGG22} considers the information-seeking scenario, where the user repeatedly asks questions and the agent answers based on the documents.
Another line of research focuses more on chit-chat scenario \citep{DBLP:conf/emnlp/MogheABK18, DBLP:conf/iclr/DinanRSFAW19}, where participants freely talk about specific topics with knowledge from the documents.
For real-world applications, both scenarios should be contained naturally.

To bridge these gaps, we propose a novel task named Proactive News Grounded Conversation, which enables dialogue systems to proactively talk about news with humans in a realistic manner.
Furthermore, we collect a human-to-human Chinese dialogue dataset \ts{NewsDialogues}, which consists of 1K conversations with 14.6K utterances and rich annotations.  
We include both information-seeking and chit-chat scenarios realistically, and an example is presented in Figure \ref{fig:example}.
To explicitly model the proactivity, we first annotate the key topics of the news article to summarize the main content of it.
Then, the agent can actively lead the conversation based on these topics, as the 1st and 7th utterances in Figure \ref{fig:example}. 
In addition, we carefully annotate the grounded knowledge of each agent utterance, including the target topic and knowledge spans, for a more informative conversation.
The major differences between our \ts{NewsDialogues} and other document-grounded dialogue datasets are summarized in Table \ref{tab:dataset_comparison}.

To further solve the problem, we propose a simple yet effective method Predict-Generate-Rank, which consists of a generator for grounded knowledge prediction and response generation, and a ranker for the ranking of multiple candidate responses to alleviate the exposure bias problem \citep{DBLP:conf/acl/ZhangFMYL19, DBLP:journals/corr/abs-2205-14690}.
We conduct comprehensive experiments based on the state-of-the-art pre-trained language models and dialogue models. 
Both automatic and human evaluation indicates that our method has substantial improvements over several baselines on \ts{NewsDialogues}.
Finally, we analyze the major limitations of current models to facilitate future research.

The main contributions are as follows.
\begin{itemize}
    \item We propose a novel task named Proactive News Grounded Conversation, aiming to empower dialogue systems to proactively talk about news with humans.
    \item To further develop this task, we build \ts{NewsDialogues}, which consists of 1K dialogues with 14.6K utterances and rich annotations.
    \item Based on \ts{NewsDialogues}, we propose a method named Predict-Generate-Rank and conduct comprehensive experiments. The results have shown the great performance of our method.
\end{itemize}


\section{Related Work}
\paragraph{Document-Grounded Conversation.}
A growing area of research is augmenting dialogue systems with external documents.
One line of research focuses on the chit-chat scenario.
\citet{DBLP:conf/emnlp/ZhouPB18, DBLP:conf/emnlp/MogheABK18} propose movie grounded conversation, where two participants talk about movies in depth based on related documents.
\ti{Wizard of Wikipedia} \citep{DBLP:conf/iclr/DinanRSFAW19} introduces more topics for conversations, totally 1,365 from Wikipedia articles.
To utilize continually updating knowledge, \citet{DBLP:conf/acl/Komeili0W22} propose \ti{Wizard of the Internet}, where dialogue systems can flexibly search relevant knowledge from the internet.

Another line of research focuses on the information-seeking scenario, where dialogue systems help users gather information through conversations \cite{DBLP:conf/emnlp/ChoiHIYYCLZ18, DBLP:journals/tacl/ReddyCM19, DBLP:conf/acl/CamposOSDCA20, DBLP:conf/sigir/Qu0CQCI20}.
Different from traditional question answering systems, the conversation context empowers dialogue systems to address open-ended and exploratory questions that need discussions to explore in depth \cite{DBLP:conf/icml/DaiCZARGG22}.
To pursue more interactive,
\citet{DBLP:conf/emnlp/FengWGPJL20, DBLP:conf/akbc/GuoZRA21, DBLP:journals/corr/abs-2207-00746} introduce clarification questions, which means that agents can also ask questions when user queries are defined as under-specified. 
Though helpful for information-seeking needs, these dialogue systems lack chatting ability.

We propose news grounded conversation, which has been neglected in previous research but is indispensable in our daily conversations.
In addition, both chit-chat and information-seeking scenarios are considered realistically.

\paragraph{Proactive Dialogue System.}
The proactivity of dialogue systems has been an open challenge. 
Previous researches model proactive topic transitions based on well-designed knowledge graphs (KGs) \citep{DBLP:conf/acl/WuGZWZLW19, DBLP:conf/acl/LiuWNWCL20}. 
However, KGs are hard to construct and have limited coverage of real-world knowledge \cite{DBLP:conf/akbc/RazniewskiSN16}. 
To explore the topic connections, \citet{DBLP:conf/acl/SevegnaniHKR20} propose the one-turn topic transition task and collect a crowdsourced dataset \ti{OTTers}. 
More recently, \citet{DBLP:conf/naacl/Cai0LY0J22} 
use reinforced self-play to train a teacher bot, which can actively 
convey knowledge during the conversation. 
However, they encourage token overlap between the generated responses and the grounded documents rather than proactive topic transition. 

We propose proactive dialogue generation based on news articles rather than KGs.
Specifically, we aim to empower dialogue systems to lead the conversation based on some key topics of the news.
To this end, we build \ts{NewsDialogues}, including 1K multi-turn dialogues.



\section{Proactive News Grounded Conversation}
\label{sec:proactive}
\label{sec:task_definition}
We propose a novel task named Proactive News Grounded Conversation.
As shown in Figure \ref{fig:example}, a user converses with an agent based on a given news article in each conversation.
The conversation begins with the agent, and during the conversation:
\begin{itemize}
    \item \textbf{User} is curious about the news and eager to chat. They can freely ask questions or express their opinions and feelings.
    \item \textbf{Agent} plays the role of a knowledgeable expert. 
    They not only reply to users in a passive way but also proactively lead the conversations based on the key topics of the news.
\end{itemize}
Following \citet{DBLP:conf/emnlp/ChoiHIYYCLZ18, DBLP:conf/iclr/DinanRSFAW19, DBLP:journals/corr/abs-2205-12609},
we introduce an information-asymmetric setting, where only the agent has access to the news article, and the user is eager to know through the conversation.
Therefore, the conversation is more open-ended and exploratory, and the agent is more helpful in real-world applications. 
Both chit-chat and information-seeking scenarios are contained naturally.

\begin{table*}[thb!]

    \centering
    \small
    \begin{tabular}{m{0.1cm}|m{1.3cm}|m{5.65cm}|m{7.0cm}}
    \toprule
        \multicolumn{1}{c|}{\tb{\#}} &
        \multicolumn{1}{c|}{\tb{Dialog Act}} & \multicolumn{1}{c|}{\tb{User Utterance}} & \multicolumn{1}{c}{\tb{Agent Utterance}}  \\
    \midrule 
    \midrule
        \centering 1 &
        \centering \tb{Chit-chat} 
        & It is indeed necessary to pay more attention to the elderly.
        & Yes, after all, we will all grow old. Help the old now, and someone will help us in the future. \\
    \midrule
    \centering 2 &
        \centering \tb{Chit-chat} 
        & Well, did the girl say why she went there?
        & \tc{huang}{I don't know}. Maybe the little girl is naughty and parents truly should take care of their children. \\
    \midrule
    \midrule
    \centering 3 &
        \centering \tb{Inform} 
        & What happened in the end? Was he saved?
        & Yes! He was \tc{lan}{found by a neighbor in time and saved}. \\
    \midrule
    \centering 4 &
        \centering \tb{Inform} 
        & Is the old man awake now? 
        & He is \tc{lan}{still in the ICU}, it is \tc{huang}{not clear} how is it going, I hope he can recover soon. \\
    \midrule
    \centering 5 &
        \centering \tb{Inform} 
        & He is so talented and loving!
        & Yeah, what he hopes most is to \tc{lan}{break the gap and barrier} between communities and people in the lockdown. \\
    \midrule
    \midrule
    \centering 6 &
        \centering \tb{Guide} 
        & -
        & 
        \tb{\ti{Topic: A police takes a choking girl to hospital.}}
        
        \tc{hong}{Have you heard} the news about a police taking a choking girl to hospital? It's so touching! 
        \\
    \midrule
    \centering 7 &
        \centering \tb{Guide} 
        & She is a genius! Maybe she can go to the Olympics after the training!
        & 
        \tb{\ti{Topic: Inherits good genes from her mother.}}
        
        It is possible! \tc{hong}{I heard that} her mother is a physical education teacher, she inherits the good genes and also develops a habit of exercising. \\
    \midrule
    \centering 8 &
        \centering \tb{Guide} 
        & So, why did this guy drive after overdosing?
        & 
        \tb{\ti{Topic: Hidden reactions of driving after overdosing.}}
        
        \tc{huang}{Not mentioned} in the news, probably he did not understand the harm of driving after overdosing. \tc{hong}{People often ignore} the adverse reactions, but they are very damaging! \\
    \midrule
    \centering 9 &
        \centering \tb{Guide} 
        & I see. Are they from an institution? Why so many people?
        & 
        \tb{\ti{Topic: 7 million yuan are swindled.}}
        
        It is a \tc{lan}{fraud gang} with many collaborators! \tc{hong}{When arrested by the police}, they had more than 180 mobile phones and swindled more than 7 million yuan. \\
    \bottomrule
    \end{tabular}
    
    \caption{
        Examples of different dialog acts of the agent. 
        We highlight some key words of \tc{lan}{inform}, \tc{hong}{guide} and answer for \tc{huang}{unanswerable} question, more details in Section \ref{sec:agent_annotator} and \ref{sec:unanswerable_question}. We also present the target topic for guide.
        For reading convenience, we translate the original Chinese to English and omit the dialog history and knowledge spans.
    }
    \label{tab:examples}
    \vspace{-5pt}
\end{table*}
\vsd

\section{\ts{NewsDialogues}}
\label{data_collection}
To further develop this task, we collect a human-to-human Chinese dialogue dataset \ts{NewsDialogues}. 

\subsection{News Article Collection}
\label{sec:news_collection}
We manually collect news articles from Toutiao\footnote{\url{https://www.toutiao.com/}, we discuss the usage policy in Section \ref{sec:term_of_use}.}, a famous news website in China.
The criteria for selection are: (1) We prefer hot news, and thus humans are more eager to talk about it.
To this end, we select news articles from the hot list in Toutiao;
(2) We only collect news articles that do not rely on image information and leave the multi-modality features for future work.


\subsection{Dialogue Collection}
\label{sec:dialogue_collection}
In \ts{NewsDialogues}, each dialogue derives from a real conversation between two human annotators, one as the user and the other one as the agent.
The conversation scenario is based on the task definition in Section \ref{sec:task_definition}, and the annotation processes for user and agent annotators are as follows.

\subsubsection{User Annotator}
\paragraph{Utterance Generation.}
User annotators freely ask questions or express their opinions and feelings.
To further investigate the behavior, we also ask them to annotate the dialog acts \cite{DBLP:conf/lrec/BuntACCFHLPPRST10} of their utterances, which are either \textbf{Question} or \textbf{Chit-chat}.
Here, chit-chat represents the comments or feelings of users, e.g., \ti{He is so talented and loving!}.

\subsubsection{Agent Annotator}
\label{sec:agent_annotator}

\paragraph{News Understanding.}
Before the conversation, the agent annotators carefully read the news articles to understand the overview. 
Then, we ask them to write the key topics of each news article, typically 2-5 short sentences.
They can write key topics in their own words or make appropriate modifications to the section titles of the news articles.



\paragraph{Utterance Generation.} 
During the conversation, the agent annotators choose appropriate dialog acts for each utterance. 
We introduce three acts, and examples are shown in Table \ref{tab:examples}.

\begin{itemize}
    \item \textbf{Chit-chat.} Naturally chat with the user without news information.
    \item \textbf{Inform.} 
    Passively respond to the user based on the knowledge from the news article. This act is appropriate when the agent answers user questions or replies to user chit-chat utterances with related news information, as the fourth and fifth examples in Table \ref{tab:examples}.
    \item \textbf{Guide.} 
    Proactively guide the current conversation based on the key topics and knowledge from the news. 
    According to our analysis, this act is appropriate under the following scenarios:
    (1) At the dialogue beginning, as the sixth case in Table \ref{tab:examples};
    (2) The current conversation is relevant to a key topic, and the agent can naturally steer the conversation to the topic, as the seventh example in Table \ref{tab:examples};
    (3) When the user asks an unanswerable question, the agent can lead the conversation to a relevant key topic, as the eighth case in Table \ref{tab:examples}. Details of unanswerable questions are given below. 
\end{itemize}
Furthermore, we find that almost $10\%$ agent utterances first inform relevant news information and then proactively lead the conversation. 
We annotate these utterances with the guide act, and an example is the last case in Table \ref{tab:examples}. 

\paragraph{Knowledge Grounding.}
When the act is inform or guide, the agent annotator can choose appropriate text spans from the news article and use them to craft a natural and informative utterance.
We annotate these spans at sentence-level, and each sentence is called a knowledge span.
Additionally, we annotate the target topic when the act is guide.
These annotations are beneficial for modularized dialogue generation \cite{DBLP:conf/acl/Zhou0HKP0LH22, DBLP:journals/corr/abs-2203-13224}, which has shown great improvements in knowledge utilization.

\subsubsection{Unanswerable Questions}
\label{sec:unanswerable_question}
During the annotation process, we find a large number of unanswerable questions, which means that there is no direct answer in the news. 
This phenomenon is common in realistic information-seeking scenarios, because human questions are open-ended and exploratory.
Most existing conversational question answering work simply replies to these questions with \textsc{No Answer} \cite{DBLP:conf/emnlp/ChoiHIYYCLZ18, DBLP:journals/tacl/ReddyCM19, DBLP:journals/tacl/AdlakhaDSVR22}.
In this paper, we adopt three strategies in order.
\begin{itemize}
    \item \textbf{Inform Relevant Information.} When there is no direct answer, but providing relevant information possibly fulfills user needs \cite{DBLP:journals/corr/abs-2207-00746}, as the fourth example in Table \ref{tab:examples}.
    \item \textbf{Guide Topic Proactively.} When there is no relevant information, but the agent can naturally steer the conversation to a relevant key topic, as the eighth case in Table \ref{tab:examples}.
    \item \textbf{Chit-chat.} When the above strategies are not suitable under the dialogue context, the agent chats with the user, as the second in Table \ref{tab:examples}.
\end{itemize}

\begin{table}[]
\small
\centering
\label{main_results}

\resizebox{\linewidth}{!}{
\begin{tabular}{@{}lrr@{}}
\toprule
\textbf{Categories} & \textbf{Statistics} & \textbf{Proportion}  \\
\midrule
\midrule
\multicolumn{3}{c}{\ti{News Article}}\\
\midrule
Total                 & 1000   &    -         \\
Avg. key topics                           & 3.44        &    -      \\
Avg. length                       & 1289.67             &  - \\
\midrule
\midrule
\multicolumn{3}{c}{\ti{Dialogues}}\\
\midrule
Total         & 1000          &   -   \\
Avg. turns                                & 14.59       &    -       \\
Avg. length of user utterances            & 17.44        &   -       \\
Avg. length of agent utterances           & 47.28      &   -          \\ \midrule
\midrule
\multicolumn{3}{c}{\ti{User Dialog Acts}}\\
\midrule
Chit-chat          & 2449  &  35.8\%                \\
Question                                & 4398         &   64.2\%       \\
Overall                                 & 6847         &   100.0\%      \\
\midrule
\midrule
\multicolumn{3}{c}{\ti{Agent Dialog Acts}}\\
\midrule
Chit-chat         & 886    &    11.4\%           \\
Guide                                    & 2876         &   37.1\%       \\
Inform                                   & 3982         &   51.4\%       \\
Overall                                  & 7744         &  100.0\%         \\
\midrule
\midrule
\multicolumn{3}{c}{\ti{Strategies for Unanswerable Questions}}\\
\midrule
Chit-chat                          & 118        &     11.2\%       \\
Guide Topic Proactively                  & 450          &   42.6\%       \\
Inform Relevant Information             & 489          &   46.3\%       \\
Overall                                   & 1057        &   100.0\%        \\
\bottomrule
\end{tabular}
}
\caption{Statistics of \ts{NewsDialogues}.}
\label{data_statistics}
\end{table}
\vspace{-10pt}

\subsection{Statistics}
The statistics of \ts{NewsDialogues} are shown in Table \ref{data_statistics}, and there are several noticeable features.
First, the news article is long and brings a new challenge to dialogue system research.
Second, as shown by the statistics of user dialog acts,
both information-seeking and chit-chat scenarios are common in \ts{NewsDialogues}.
The large proportion of user questions ($64.2\%$) indicates that information-seeking scenario is indispensable for real-world applications.
Third, unanswerable questions occupy a large proportion of user questions (1057 of 4398). 
Therefore, it is important for dialogue systems to address these questions properly.
\section{Method}
\subsection{Task formulation}
Each conversation is grounded on a news article \tb{\ti{n}} with key topics \tb{\ti{k}}, and the dialogue system learns to generate a response \tb{\ti{r}} based on the dialog history \tb{\ti{d}}.
In addition, it should also predict the grounded knowledge \tb{\ti{g}}, including both the target topic and the knowledge spans for generation, when needed. 


\subsection{Predict-Generate-Rank}
\begin{figure}
    \centering
    \includegraphics[width=\linewidth]{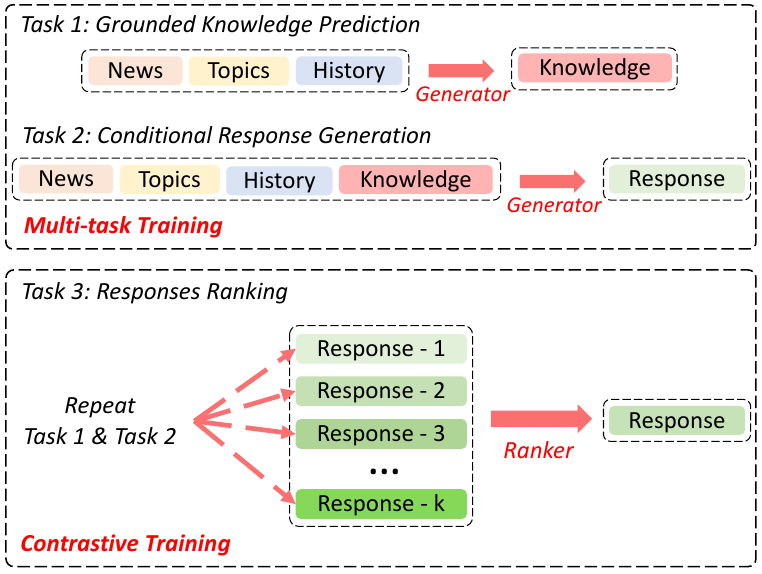}
    \caption{The overview of our Predict-Generate-Rank, including a generator trained with a multi-task objective and a ranker trained with contrastive loss.
    }
    \label{fig:method}
\end{figure}

We propose a simple yet effective method, named Predict-Generate-Rank, including a three stage generation process, as shown in Figure \ref{fig:method}.

\paragraph{Task 1: Grounded Knowledge Prediction.} 
The model first predicts the grounded knowledge $\tb{\ti{g}}$ for response generation.
Specifically, we concatenate the target topic and the knowledge spans as $\tb{\ti{g}}$\footnote{Both the target topic and the knowledge spans can be none, depending on the dialog act. When they are none, $\tb{\ti{g}}$ is an empty string.}, and formulate this problem as a task of language generation.
The objective is the negative log-likelihood:
\begin{equation}
   \mathcal{L}_1 = - \sum_{l=1}^L \log P(\ti{g}_l|\ti{g}_{<l}, \tb{\ti{n}}, \tb{\ti{k}}, \tb{\ti{d}}), \nonumber
\end{equation}
where $\ti{g}_l$ represents the $l$-th token of $\tb{\ti{g}}$, and $L$ is the total length.

\paragraph{Task 2: Conditional Response Generation.}
Based on the grounded knowledge $\tb{\ti{g}}$, our model learns to generate the response autoregressively. 
The objective function is as follows:
\begin{equation}
   \mathcal{L}_2 = - \sum_{t=1}^T \log P(\ti{r}_t|\ti{r}_{<t}, \tb{\ti{n}}, \tb{\ti{k}}, \tb{\ti{d}}, \tb{\ti{g}}), \nonumber
\end{equation}
where $\ti{r}_t$ denotes the $t$-th token of $\tb{\ti{r}}$, and $T$ is the total length.
We use the ground-truth knowledge for training and the predicted knowledge for inference. 
Our generator is trained with a multi-task objective: $\mathcal{L} = \mathcal{L}_1 + \mathcal{L}_2$, as in \citet{DBLP:journals/tacl/PengLLSLG21}.

\paragraph{Task 3: Responses Ranking.}
One major problem of the above tasks is the gap between the ground-truth knowledge and the predicted knowledge, which results in severe exposure bias \citep{DBLP:conf/acl/ZhangFMYL19, DBLP:journals/corr/abs-2205-14690} for text generation.
Particularly, the generated response can be low-quality if the predicted knowledge is irrelevant to the dialogue context. 
To alleviate this problem, we further introduce a ranking task.
Specifically, the generator first samples multiple knowledge and generates the responses based on them. Then, a ranker is used to select the best response.

We use a simple strategy to construct datasets for the training of the ranker.
First, we finetune the generator on the training set of \ts{NewsDialogues}, then we use this model to sample knowledge and generate responses on the training set, and we can get $D={\{(\hat{\tb{\ti{g}}}_m, \hat{\tb{\ti{r}}}_m)\}}_{m=1}^M$ for each example, where $\hat{\tb{\ti{g}}}$ is the predicted knowledge and $\hat{\tb{\ti{r}}}$ is the response conditioned on $\hat{\tb{\ti{g}}}$.
For each $(\hat{\tb{\ti{g}}}, \hat{\tb{\ti{r}}})$, we compute the matching scores with the ground truth $(\tb{\ti{g}}, \tb{\ti{r}})$:
\begin{equation}
\begin{aligned}
    \Delta_1 (\tb{\ti{g}}, \hat{\tb{\ti{g}}}) &= \tb{Word-Level F1} (\tb{\ti{g}}, \hat{\tb{\ti{g}}}), \\
    \Delta_2 (\tb{\ti{r}}, \hat{\tb{\ti{r}}}) &= \text{\tb{BLEU}-4}  (\tb{\ti{r}}, \hat{\tb{\ti{r}}}). \nonumber   
\end{aligned}
\end{equation}
The responses with $\Delta_1 > \gamma_1$ and $\Delta_2 > \gamma_2$ are set as positive examples, which means both the knowledge and responses are similar to the ground truths, while other responses are set as negative examples. 
Then, we can get the training set for the ranker, and the validation set for the ranker is constructed with the same strategy on the validation set of \ts{NewsDialogues}.

Suppose $\mathcal{\hat{R}}^+$ is the set of positive examples and $\mathcal{\hat{R}}^-$ is the set of negative examples, we train the ranker with contrastive loss: 
\begin{equation}
    \mathcal{L}_3 = - \sum_{\hat{\tb{\ti{r}}}^+ \in \mathcal{\hat{R}}^+} \log \frac{\text{exp}^{s_{\hat{\tb{\ti{r}}}^+}}}{\text{exp}^{s_{\hat{\tb{\ti{r}}}^+}} + \sum_{\hat{\tb{\ti{r}}}^- \in \mathcal{\hat{R}}^-}\text{exp}^{s_{\hat{\tb{\ti{r}}}^-}}}, \nonumber
\end{equation}
where $s_{\hat{\tb{\ti{r}}}} = D_\phi([\tb{\ti{d}}, \hat{\tb{\ti{r}}}])$ is the ranker score and $D_\phi$ represents the ranker, which is BERT \citep{DBLP:conf/naacl/DevlinCLT19} in this paper. 
The input is the concatenation of the dialogue history $\tb{\ti{d}}$ and the response $\hat{\tb{\ti{r}}}$, and $s_{\hat{\tb{\ti{r}}}} \in \mathcal{R}$ is computed by the representation of \cls token and a linear projection layer.
We pre-train the ranker on DuConv \citep{DBLP:conf/acl/WuGZWZLW19} and KdConv \citep{DBLP:conf/acl/ZhouZHHZ20} to better capture the relation between dialogue histories and responses, more details are given in Appendix \ref{app:implementation_details}.

\paragraph{Inference.} For inference, the generator first samples $k$ grounded knowledge and generates responses based on them. Then, we use the ranker to select the response with the highest score.

\begin{table*}[t]
    \begin{center}
    \centering
    \small
    \resizebox{\linewidth}{!}{
    \begin{tabular}{lccccccccc}
    \toprule
       \tb{Model} & \tb{Topic F1} & \tb{Span F1} & \tb{BLEU-1} & \tb{BLEU-2} & \tb{BLEU-4} & \tb{ROUGE-2} & \tb{ROUGE-L} & $\tb{Distinct-2}^*$ \\
    \midrule
    \midrule
        \multicolumn{9}{c}{\it{Dialogue Model}}\\
    \midrule
        CDial-GPT \citep{wang2020chinese} & - & - & 14.22 & 4.56 & 0.27 & 2.83 & 13.32 & 47.65 \\
        EVA2.0 \citep{DBLP:journals/corr/abs-2203-09313} & - & - & 13.72 & 3.56 & 0.14 & 2.11 & 13.35 & \tb{50.57} \\
    \midrule
    \midrule
        \multicolumn{9}{c}{\it{End-to-end Model}}\\
    \midrule
        BLOOM \citep{DBLP:journals/corr/abs-2211-05100} & 41.00 & 28.42 & 18.78 & 9.23 & 3.23 & 7.34 & 19.18 & 46.47 \\
        mBART \citep{DBLP:journals/corr/abs-2008-00401} & 13.47 & 28.74 & 14.33 & 8.42 & 4.21 & 7.83 & 17.33 & 37.60 \\
        mT5 \citep{xue2020mt5} & 35.49 & 27.94 & 17.18 & 8.46 & 3.35 & 7.06 & 18.40 & 47.05 \\
        Chinese GPT \citep{DBLP:conf/emnlp/ZhaoCZZLLCDJD19} & 37.37 & 23.96 & 16.36 & 7.58 & 2.28 & 5.67 & 16.59 & 37.50 \\
        Chinese BART \citep{fengshenbang} & 13.01 & 26.73 & 15.78 & 7.08 & 1.39 & 5.82 & 18.95 & 39.49 \\
        Chinese T5 \citep{fengshenbang} & 41.92 & 39.66 & 25.42 & 16.03 & 8.41 & 13.92 & 26.33 & 45.41 \\
    \midrule
    \midrule
        Predict-Generate-Rank & \tb{43.03} & \tb{43.35} & \tb{28.88} & \tb{19.47} & \tb{10.99} & \tb{17.41} & \tb{29.98} & 42.45 \\
    \midrule
    \midrule
        Human & 100.00 & 100.00 & 100.00 & 100.00 & 100.00 & 100.00 & 100.00 & 51.06 \\
    \bottomrule
    \end{tabular}
    }
    \end{center}

    \caption{
        Automatic evaluation on the test set of \ts{NewsDialogues}.
        All metrics evaluate the relevance between generations and ground truths except Distinct-2.
        We list Distinct-2 for the reference of diversity, which is the proportion of distinct bigrams in the total generations and has no relation with the ground truths.
    }
    \label{tab:main_results}
    \vspace{-5pt}
\end{table*}

\section{Experiments}

\subsection{Baselines}

\paragraph{Dialogue Model.} We first investigate the performance of dialogue models. Specifically, we finetune the models on \ts{NewsDialogues} with only dialogue data, the input is the dialogue history and the target is the ground-truth response.
As \ts{NewsDialogues} is based on Chinese, we evaluate the performance of Chinese dialogue models, CDial-GPT \citep{wang2020chinese} and EVA2.0 \citep{DBLP:journals/corr/abs-2203-09313}.
EVA2.0 has shown the state-of-the-art performance on Chinese dialogue generation.

\paragraph{End-to-end Model.} 
We finetune the pre-trained language models to predict the grounded knowledge and generate the response based on it sequentially.
The training process is the same as our prediction and generation task with a multi-task objective. 
We evaluate a series of models, including BLOOM \citep{DBLP:journals/corr/abs-2211-05100} (Multilingual GPT), mBART \citep{DBLP:journals/corr/abs-2008-00401} (Multilingual BART), mT5 \citep{xue2020mt5} (Multilingual T5), Chinese GPT \citep{DBLP:conf/emnlp/ZhaoCZZLLCDJD19}, Chinese BART \citep{fengshenbang} and Chinese T5 \citep{fengshenbang}.

\subsection{Implementation}
We randomly split \ts{NewsDialogues} into the train / validation / test sets with a ratio of $8:1:1$, and the numbers of dialogues are 800, 100, and 100.
For our Predict-Generate-Rank model, we use Chinese T5 as the generator \citep{fengshenbang} and Mengzi-Bert-base (Chinese BERT) \citep{zhang2021mengzi} as the ranker. 
The $\gamma_1$ and $\gamma_2$ are set as 50 and 15, and the candidate num $k$ is set as 16.
More details are shown in Appendix  \ref{app:implementation_details}.

\subsection{Automatic Evaluation}
\ntb{Metrics.}
We adopt BLEU \cite{DBLP:conf/acl/PapineniRWZ02}, ROUGE \cite{lin-2004-rouge} and Distinct \cite{DBLP:conf/naacl/LiGBGD16} for the evaluation of response generation. 
In addition, we compute Topic F1 score to evaluate topic prediction and word-level F1 score for knowledge span prediction (Span F1) as in \citet{DBLP:conf/emnlp/ChoiHIYYCLZ18}. 

\ntb{Results.}
As shown in Table \ref{tab:main_results}, dialogue models perform less competitively than other models. 
The reason stems from the lack of news information, which is indispensable for \ts{NewsDialogues}.
In addition, dialogue models show the best diversity, and we conjecture this benefits from the pre-training with large-scale conversation data, which contains abundant topics.
For end-to-end models, BART performs poorly as it uses absolute position embedding with the maximum length of 1024, which is not sufficient when the news article is long.
T5 models with relative position embedding and BLOOM with the maximum length of 2048 can alleviate this problem.
The proposed Predict-Generate-Rank improves the performance substantially, except for diversity. 
We focus more on the relevance between predicted responses and ground-truth responses, which is reflected by other metrics.



\subsection{Human Interactive Evaluation}
To investigate the performance more realistically, we employ human annotators to converse with different models, humans acting as users while models acting as agents. 
As human interactive evaluation has a high cost, we only evaluate the best end-to-end model Chinese T5 and our Predict-Generate-Rank.
More details are shown in Appendix \ref{app:human_interactive_evaluation_setting}.

\ntb{Metrics.}
(1) \ti{Fluency}: whether the response is fluent and understandable.
(2) \ti{Coherence}: whether the response is coherent and consistent with the context.
(3) \ti{Naturalness}: If the response has a target topic, is the topic transition natural and appropriate?
(4) \ti{Knowledgeability}: whether the agent is knowledgeable of the news and uses knowledge reasonably. 
(5) \ti{Proactivity}: whether the agent is proactive and helps you understand the content of the news.
(6) \ti{Engagingness}: whether the conversation is engaging and gives you a happy surprise.
The first three metrics are utterance-level, while others are dialogue-level.
Each score is on a scale from $1$ to $3$, meaning bad, moderate, and good.

\begin{table}[t]
    \begin{center}
    \centering
    \small
    \resizebox{\linewidth}{!}{%
    \begin{tabular}{lcccccc}
    \toprule
       \tb{Model} & \tb{Flu.} & \tb{Coh.} & \tb{Nat.} & \tb{Kno.} & \tb{Pro.} & \tb{Eng.}\\
    \midrule
    \midrule
        Chinese-T5 & 2.51 & 1.94 & 2.09 & 1.91 & 1.66 & 1.63\\
        Predict-Generate-Rank & \tb{2.53} & \tb{1.99} & \tb{2.16} & \tb{2.15} & \tb{1.92} & \tb{1.71}\\
    \midrule
    \midrule
        Human & 2.97 & 2.91 & 2.60 & 2.95 & 2.80 & 2.70\\
    \bottomrule
    \end{tabular}
    }
    \end{center}

    \caption{
        Human Interactive Evaluation on \ts{NewsDialogues}, where Flu., Coh., Nat., Kno., Pro. and Eng. represent \ti{Fluency}, \ti{Coherence}, \ti{Naturalness}, \ti{Knowledgeability}, \ti{Proactivity} and \ti{Engagingness} respectively.
    }
    \label{tab:human_interactive_evaluation}
    \vspace{-5pt}
\end{table}

\begin{table}[t]
    \begin{center}
    \centering
    \small
    \resizebox{\linewidth}{!}{%
    \begin{tabular}{ccccc}
    \toprule
       \tb{Number} & \tb{Topic F1} & \tb{Span F1} & \tb{BLEU-4} & \tb{ROUGE-L} \\
    \midrule
    \midrule
        $k=1$ & 41.92 & 39.66 & 8.41 &26.33 \\
        $k=4$ & 42.13 & 43.44 & 10.33 & 29.00 \\
        $k=8$ & 41.86 & \tb{43.56} & 10.73 & 29.81 \\
        $k=16$ & \tb{43.03} & 43.35 & \tb{10.99} & \tb{29.98}\\
        $k=24$ & 42.61 & 43.13 & 10.76 & 29.17 \\
    \bottomrule
    \end{tabular}
    }
    \end{center}

    \caption{
        Analysis studies on the candidates number $k$ of Predict-Generate-Rank.
    }
    \label{tab:impact_of_k}
    \vspace{-5pt}
\end{table}

\ntb{Results.}
As shown in Table \ref{tab:human_interactive_evaluation},
two models show comparable fluency and coherence, and both are far from perfect.
For the naturalness of topic transition, Predict-Generate-Rank performs slightly better.
Surprisingly, the human score is only $2.60$, which indicates the challenge of natural topic transition.
Regarding the dialogue-level metrics, our model greatly improves knowledgeability and proactivity, which is consistent with the better performance on topic and knowledge span prediction in automatic evaluation. 
Furthermore, human evaluators feel more engaged when talking with Predict-Generate-Rank.
Nevertheless, there is a large gap between current models and humans in many aspects, indicating plenty of room for improvement.

\subsection{Impact of Ranking}
We conduct experiments to investigate the impact of the ranking task.
As shown in Table \ref{tab:impact_of_k}, the performance improves when more candidates are generated, and the Span F1 score has an improvement of $3.78$ when only four candidates are generated.
Our method gets the best results when $k=16$, which is the default setting in this paper.
According to our manual check, the ranker helps select more relevant responses, thus contributing to the improvements.

\subsection{Discussion}
Based on the above results, we conclude three major defects of current models.
First, they have poor conversation ability, as the low human score in \ti{fluency} and \ti{coherence}.
This problem derives from the scale of \ts{NewsDialogues}, and a possible way is using the large-scale conversation data in the general domain for pre-training.
Second, current models cannot use news knowledge appropriately, as the low Span F1 and \ti{Knowledgeability}.
According to our analysis, the reasons are in many aspects:
(1) The grounded news is typically long and complex.
(2) Many utterances are contextual, and the dialogue system needs to resolve the frequent coreference and information omission \citep{DBLP:conf/emnlp/ElgoharyPB19} for knowledge extraction. 
Considering the second utterance in Figure \ref{fig:example}, the agent needs to know that ``her'' represents the ``baby girl'' in the first utterance.
(3) Rather than answering factoid questions in most existing QA datasets, the conversation scenario is much more open-ended, and commonsense reasoning ability is necessary. As the 4th example in Table \ref{tab:examples}, only when the dialogue system knows the relation between ``awake'' and ``ICU'', can it find the knowledge for a generation.
Third, current models are incapable of natural and proactive topic transitions, as the low Topic F1, \ti{Naturalness}, and \ti{Proactivity}.
This also stems from the lack of commonsense knowledge and reasoning skills to capture the relations between current topics and relevant topics.
This is a valuable characteristic of \ts{NewsDialogues}, which is challenging but rewarding for dialogue system research.
\vsd



\section{Conclusion}
In this paper, we define a novel task named Proactive News Grounded Conversation,
where the dialogue system can proactively lead the conversation based on some topics of the news.
In addition, we collect \ts{NewsDialogues} with 1K dialogues and rich annotations.
Furthermore, we propose Predict-Generate-Rank, which consists of a generator trained with a multi-task objective and a ranker trained with contrastive loss.
Comprehensive experiments have been conducted to investigate the performance of current models on \ts{NewsDialogues}.
We hope that our research will spur the development of dialogue systems that are more proactive and knowledgeable in various scenarios. 

\section*{Limitations}
We acknowledge the following limitations of our work.

\paragraph{Limitations of \ts{NewsDialogues}.}
First, we only collect 1K human-to-human conversations with 14.6K utterances due to the high cost of the annotation process (Section \ref{sec:dialogue_collection}).
This brings difficulties for the learning of news grounded dialogue generation.
Second, each conversation in \ts{NewsDialogues} is grounded on one news article, which may have limited knowledge for real-world applications.
We leave the multi-article grounded setting for future work.
Third, as mentioned in Section \ref{sec:news_collection}, the image information in the news article is neglected in this version, which requires further exploration.

\paragraph{Limitations of Experiments.}
Large language models (LLM) have shown great few-shot learning ability and generation capacity on various tasks, e.g., GPT-3 \citep{DBLP:conf/nips/BrownMRSKDNSSAA20}, OPT-175B \citep{DBLP:journals/corr/abs-2205-01068} and BLOOM-176B \citep{DBLP:journals/corr/abs-2211-05100} etc.
It is important to investigate the performance of LLM on \ts{NewsDialogues}, while this has been neglected in this work due to the limited computational resources.
In addition, it is also valuable to investigate the performance of ChatGPT\footnote{\url{https://openai.com/blog/chatgpt/}} on \ts{NewsDialogues}, and we leave this for our future work.

\section*{Ethics Statement}
\label{sec:ethical_considerations}

\subsection*{Private Information}
We carefully remove all personal information through the data cleaning process: 
First, we do not include any account information during the data collecting procedure, which means all the data are anonymous. 
Second, we clean the potential private information such as emails, ID numbers, phone numbers, etc. in the data to further ensure the privacy.   

\subsection*{Offensive Content}
We have taken two steps to avoid offensive content in \ts{NewsDialogues}.
First, we ask the annotators not to speak offensive content during the conversations.
Second, we manually check all conversations after data collection and throw away the conversations including offensive content.

\subsection*{Terms of Use}
\label{sec:term_of_use}
Upon acceptance, we will provide all the codes and the proposed dataset \ts{NewsDialogues} including conversations, annotations for knowledge and topics, and corresponding URLs for the News according to the terms of use of Toutiao\footnote{\url{https://www.toutiao.com/user_agreement/}}.
\ts{NewsDialogues} is only used for facilitating dialogue system research and can not be used for any commercial purposes.

\section*{Acknowledgements}
This work was partly supported by the National Key Research and Development Program of China (No. 2020YFB1708200) ,  the "Graph Neural Network Project" of Ping An Technology (Shenzhen) Co., Ltd. and AMiner.Shenzhen SciBrain fund.

\bibliography{anthology,custom}

\begin{thebibliography}{50}
\expandafter\ifx\csname natexlab\endcsname\relax\def\natexlab#1{#1}\fi

\bibitem[{Adlakha et~al.(2022)Adlakha, Dhuliawala, Suleman, de~Vries, and
  Reddy}]{DBLP:journals/tacl/AdlakhaDSVR22}
Vaibhav Adlakha, Shehzaad Dhuliawala, Kaheer Suleman, Harm de~Vries, and Siva
  Reddy. 2022.
\newblock Topiocqa: Open-domain conversational question answering with topic
  switching.
\newblock \emph{Trans. Assoc. Comput. Linguistics}, 10:468--483.

\bibitem[{An et~al.(2022)An, Feng, Lv, Kong, Qiu, and
  Huang}]{DBLP:journals/corr/abs-2205-14690}
Chenxin An, Jiangtao Feng, Kai Lv, Lingpeng Kong, Xipeng Qiu, and Xuanjing
  Huang. 2022.
\newblock Cont: Contrastive neural text generation.
\newblock \emph{CoRR}, abs/2205.14690.

\bibitem[{Brown et~al.(2020)Brown, Mann, Ryder, Subbiah, Kaplan, Dhariwal,
  Neelakantan, Shyam, Sastry, Askell, Agarwal, Herbert{-}Voss, Krueger,
  Henighan, Child, Ramesh, Ziegler, Wu, Winter, Hesse, Chen, Sigler, Litwin,
  Gray, Chess, Clark, Berner, McCandlish, Radford, Sutskever, and
  Amodei}]{DBLP:conf/nips/BrownMRSKDNSSAA20}
Tom~B. Brown, Benjamin Mann, Nick Ryder, Melanie Subbiah, Jared Kaplan,
  Prafulla Dhariwal, Arvind Neelakantan, Pranav Shyam, Girish Sastry, Amanda
  Askell, Sandhini Agarwal, Ariel Herbert{-}Voss, Gretchen Krueger, Tom
  Henighan, Rewon Child, Aditya Ramesh, Daniel~M. Ziegler, Jeffrey Wu, Clemens
  Winter, Christopher Hesse, Mark Chen, Eric Sigler, Mateusz Litwin, Scott
  Gray, Benjamin Chess, Jack Clark, Christopher Berner, Sam McCandlish, Alec
  Radford, Ilya Sutskever, and Dario Amodei. 2020.
\newblock Language models are few-shot learners.
\newblock In \emph{Advances in Neural Information Processing Systems 33: Annual
  Conference on Neural Information Processing Systems 2020, NeurIPS 2020,
  December 6-12, 2020, virtual}.

\bibitem[{Bunt et~al.(2010)Bunt, Alexandersson, Carletta, Choe, Fang, Hasida,
  Lee, Petukhova, Popescu{-}Belis, Romary, Soria, and
  Traum}]{DBLP:conf/lrec/BuntACCFHLPPRST10}
Harry Bunt, Jan Alexandersson, Jean Carletta, Jae{-}Woong Choe, Alex~Chengyu
  Fang, K{\^{o}}iti Hasida, Kiyong Lee, Volha Petukhova, Andrei
  Popescu{-}Belis, Laurent Romary, Claudia Soria, and David~R. Traum. 2010.
\newblock Towards an {ISO} standard for dialogue act annotation.
\newblock In \emph{Proceedings of the International Conference on Language
  Resources and Evaluation, {LREC} 2010, 17-23 May 2010, Valletta, Malta}.
  European Language Resources Association.

\bibitem[{Cai et~al.(2022)Cai, Wan, Liu, Yu, Yu, and
  Joshi}]{DBLP:conf/naacl/Cai0LY0J22}
Pengshan Cai, Hui Wan, Fei Liu, Mo~Yu, Hong Yu, and Sachindra Joshi. 2022.
\newblock Learning as conversation: Dialogue systems reinforced for information
  acquisition.
\newblock In \emph{Proceedings of the 2022 Conference of the North American
  Chapter of the Association for Computational Linguistics: Human Language
  Technologies, {NAACL} 2022, Seattle, WA, United States, July 10-15, 2022},
  pages 4781--4796. Association for Computational Linguistics.

\bibitem[{Campos et~al.(2020)Campos, Otegi, Soroa, Deriu, Cieliebak, and
  Agirre}]{DBLP:conf/acl/CamposOSDCA20}
Jon~Ander Campos, Arantxa Otegi, Aitor Soroa, Jan Deriu, Mark Cieliebak, and
  Eneko Agirre. 2020.
\newblock Doqa - accessing domain-specific faqs via conversational {QA}.
\newblock In \emph{Proceedings of the 58th Annual Meeting of the Association
  for Computational Linguistics, {ACL} 2020, Online, July 5-10, 2020}, pages
  7302--7314. Association for Computational Linguistics.

\bibitem[{Choi et~al.(2018)Choi, He, Iyyer, Yatskar, Yih, Choi, Liang, and
  Zettlemoyer}]{DBLP:conf/emnlp/ChoiHIYYCLZ18}
Eunsol Choi, He~He, Mohit Iyyer, Mark Yatskar, Wen{-}tau Yih, Yejin Choi, Percy
  Liang, and Luke Zettlemoyer. 2018.
\newblock Quac: Question answering in context.
\newblock In \emph{Proceedings of the 2018 Conference on Empirical Methods in
  Natural Language Processing, Brussels, Belgium, October 31 - November 4,
  2018}, pages 2174--2184. Association for Computational Linguistics.

\bibitem[{Dai et~al.(2022)Dai, Chaganty, Zhao, Amini, Rashid, Green, and
  Guu}]{DBLP:conf/icml/DaiCZARGG22}
Zhuyun Dai, Arun~Tejasvi Chaganty, Vincent~Y. Zhao, Aida Amini, Qazi~Mamunur
  Rashid, Mike Green, and Kelvin Guu. 2022.
\newblock Dialog inpainting: Turning documents into dialogs.
\newblock In \emph{International Conference on Machine Learning, {ICML} 2022,
  17-23 July 2022, Baltimore, Maryland, {USA}}, volume 162 of \emph{Proceedings
  of Machine Learning Research}, pages 4558--4586. {PMLR}.

\bibitem[{Devlin et~al.(2019)Devlin, Chang, Lee, and
  Toutanova}]{DBLP:conf/naacl/DevlinCLT19}
Jacob Devlin, Ming{-}Wei Chang, Kenton Lee, and Kristina Toutanova. 2019.
\newblock {BERT:} pre-training of deep bidirectional transformers for language
  understanding.
\newblock In \emph{Proceedings of the 2019 Conference of the North American
  Chapter of the Association for Computational Linguistics: Human Language
  Technologies, {NAACL-HLT} 2019, Minneapolis, MN, USA, June 2-7, 2019, Volume
  1 (Long and Short Papers)}, pages 4171--4186. Association for Computational
  Linguistics.

\bibitem[{Dinan et~al.(2019)Dinan, Roller, Shuster, Fan, Auli, and
  Weston}]{DBLP:conf/iclr/DinanRSFAW19}
Emily Dinan, Stephen Roller, Kurt Shuster, Angela Fan, Michael Auli, and Jason
  Weston. 2019.
\newblock Wizard of wikipedia: Knowledge-powered conversational agents.
\newblock In \emph{7th International Conference on Learning Representations,
  {ICLR} 2019, New Orleans, LA, USA, May 6-9, 2019}. OpenReview.net.

\bibitem[{Elgohary et~al.(2019)Elgohary, Peskov, and
  Boyd{-}Graber}]{DBLP:conf/emnlp/ElgoharyPB19}
Ahmed Elgohary, Denis Peskov, and Jordan~L. Boyd{-}Graber. 2019.
\newblock Can you unpack that? learning to rewrite questions-in-context.
\newblock In \emph{Proceedings of the 2019 Conference on Empirical Methods in
  Natural Language Processing and the 9th International Joint Conference on
  Natural Language Processing, {EMNLP-IJCNLP} 2019, Hong Kong, China, November
  3-7, 2019}, pages 5917--5923. Association for Computational Linguistics.

\bibitem[{Feng et~al.(2020)Feng, Wan, Gunasekara, Patel, Joshi, and
  Lastras}]{DBLP:conf/emnlp/FengWGPJL20}
Song Feng, Hui Wan, R.~Chulaka Gunasekara, Siva~Sankalp Patel, Sachindra Joshi,
  and Luis~A. Lastras. 2020.
\newblock doc2dial: {A} goal-oriented document-grounded dialogue dataset.
\newblock In \emph{Proceedings of the 2020 Conference on Empirical Methods in
  Natural Language Processing, {EMNLP} 2020, Online, November 16-20, 2020},
  pages 8118--8128. Association for Computational Linguistics.

\bibitem[{Gu et~al.(2022)Gu, Wen, Sun, Song, Ke, Zheng, Zhang, Yao, Zhu, Tang,
  and Huang}]{DBLP:journals/corr/abs-2203-09313}
Yuxian Gu, Jiaxin Wen, Hao Sun, Yi~Song, Pei Ke, Chujie Zheng, Zheng Zhang,
  Jianzhu Yao, Xiaoyan Zhu, Jie Tang, and Minlie Huang. 2022.
\newblock {EVA2.0:} investigating open-domain chinese dialogue systems with
  large-scale pre-training.
\newblock \emph{CoRR}, abs/2203.09313.

\bibitem[{Guo et~al.(2021)Guo, Zhang, Reddy, and
  Alikhani}]{DBLP:conf/akbc/GuoZRA21}
Meiqi Guo, Mingda Zhang, Siva Reddy, and Malihe Alikhani. 2021.
\newblock Abg-coqa: Clarifying ambiguity in conversational question answering.
\newblock In \emph{3rd Conference on Automated Knowledge Base Construction,
  {AKBC} 2021, Virtual, October 4-8, 2021}.

\bibitem[{Holtzman et~al.(2020)Holtzman, Buys, Du, Forbes, and
  Choi}]{DBLP:conf/iclr/HoltzmanBDFC20}
Ari Holtzman, Jan Buys, Li~Du, Maxwell Forbes, and Yejin Choi. 2020.
\newblock The curious case of neural text degeneration.
\newblock In \emph{8th International Conference on Learning Representations,
  {ICLR} 2020, Addis Ababa, Ethiopia, April 26-30, 2020}. OpenReview.net.

\bibitem[{Huang et~al.(2020)Huang, Zhu, and Gao}]{huang2020challenges}
Minlie Huang, Xiaoyan Zhu, and Jianfeng Gao. 2020.
\newblock Challenges in building intelligent open-domain dialog systems.
\newblock \emph{ACM Transactions on Information Systems (TOIS)}, 38(3):1--32.

\bibitem[{Kim et~al.(2022)Kim, Kim, Yoo, and
  Kang}]{DBLP:journals/corr/abs-2205-12609}
Gangwoo Kim, Sungdong Kim, Kang~Min Yoo, and Jaewoo Kang. 2022.
\newblock Towards more realistic generation of information-seeking
  conversations.
\newblock \emph{CoRR}, abs/2205.12609.

\bibitem[{Kingma and Ba(2015)}]{DBLP:journals/corr/KingmaB14}
Diederik~P. Kingma and Jimmy Ba. 2015.
\newblock Adam: {A} method for stochastic optimization.
\newblock In \emph{3rd International Conference on Learning Representations,
  {ICLR} 2015, San Diego, CA, USA, May 7-9, 2015, Conference Track
  Proceedings}.

\bibitem[{Komeili et~al.(2022)Komeili, Shuster, and
  Weston}]{DBLP:conf/acl/Komeili0W22}
Mojtaba Komeili, Kurt Shuster, and Jason Weston. 2022.
\newblock Internet-augmented dialogue generation.
\newblock In \emph{Proceedings of the 60th Annual Meeting of the Association
  for Computational Linguistics (Volume 1: Long Papers), {ACL} 2022, Dublin,
  Ireland, May 22-27, 2022}, pages 8460--8478. Association for Computational
  Linguistics.

\bibitem[{Li et~al.(2016)Li, Galley, Brockett, Gao, and
  Dolan}]{DBLP:conf/naacl/LiGBGD16}
Jiwei Li, Michel Galley, Chris Brockett, Jianfeng Gao, and Bill Dolan. 2016.
\newblock A diversity-promoting objective function for neural conversation
  models.
\newblock In \emph{{NAACL} {HLT} 2016, The 2016 Conference of the North
  American Chapter of the Association for Computational Linguistics: Human
  Language Technologies, San Diego California, USA, June 12-17, 2016}, pages
  110--119. The Association for Computational Linguistics.

\bibitem[{Lin(2004)}]{lin-2004-rouge}
Chin-Yew Lin. 2004.
\newblock \href {https://aclanthology.org/W04-1013} {{ROUGE}: A package for
  automatic evaluation of summaries}.
\newblock In \emph{Text Summarization Branches Out}, pages 74--81, Barcelona,
  Spain. Association for Computational Linguistics.

\bibitem[{Liu et~al.(2020)Liu, Wang, Niu, Wu, Che, and
  Liu}]{DBLP:conf/acl/LiuWNWCL20}
Zeming Liu, Haifeng Wang, Zheng{-}Yu Niu, Hua Wu, Wanxiang Che, and Ting Liu.
  2020.
\newblock Towards conversational recommendation over multi-type dialogs.
\newblock In \emph{Proceedings of the 58th Annual Meeting of the Association
  for Computational Linguistics, {ACL} 2020, Online, July 5-10, 2020}, pages
  1036--1049. Association for Computational Linguistics.

\bibitem[{Moghe et~al.(2018)Moghe, Arora, Banerjee, and
  Khapra}]{DBLP:conf/emnlp/MogheABK18}
Nikita Moghe, Siddhartha Arora, Suman Banerjee, and Mitesh~M. Khapra. 2018.
\newblock Towards exploiting background knowledge for building conversation
  systems.
\newblock In \emph{Proceedings of the 2018 Conference on Empirical Methods in
  Natural Language Processing, Brussels, Belgium, October 31 - November 4,
  2018}, pages 2322--2332. Association for Computational Linguistics.

\bibitem[{Ni et~al.(2021)Ni, Young, Pandelea, Xue, Adiga, and
  Cambria}]{ni2021recent}
Jinjie Ni, Tom Young, Vlad Pandelea, Fuzhao Xue, Vinay Adiga, and Erik Cambria.
  2021.
\newblock Recent advances in deep learning based dialogue systems: A systematic
  survey.
\newblock \emph{arXiv preprint arXiv:2105.04387}.

\bibitem[{Papineni et~al.(2002)Papineni, Roukos, Ward, and
  Zhu}]{DBLP:conf/acl/PapineniRWZ02}
Kishore Papineni, Salim Roukos, Todd Ward, and Wei{-}Jing Zhu. 2002.
\newblock Bleu: a method for automatic evaluation of machine translation.
\newblock In \emph{Proceedings of the 40th Annual Meeting of the Association
  for Computational Linguistics, July 6-12, 2002, Philadelphia, PA, {USA}},
  pages 311--318. {ACL}.

\bibitem[{Peng et~al.(2021)Peng, Li, Li, Shayandeh, Liden, and
  Gao}]{DBLP:journals/tacl/PengLLSLG21}
Baolin Peng, Chunyuan Li, Jinchao Li, Shahin Shayandeh, Lars Liden, and
  Jianfeng Gao. 2021.
\newblock {SOLOIST:} building task bots at scale with transfer learning and
  machine teaching.
\newblock \emph{Trans. Assoc. Comput. Linguistics}, 9:907--824.

\bibitem[{Qu et~al.(2020)Qu, Yang, Chen, Qiu, Croft, and
  Iyyer}]{DBLP:conf/sigir/Qu0CQCI20}
Chen Qu, Liu Yang, Cen Chen, Minghui Qiu, W.~Bruce Croft, and Mohit Iyyer.
  2020.
\newblock Open-retrieval conversational question answering.
\newblock In \emph{Proceedings of the 43rd International {ACM} {SIGIR}
  conference on research and development in Information Retrieval, {SIGIR}
  2020, Virtual Event, China, July 25-30, 2020}, pages 539--548. {ACM}.

\bibitem[{Rasley et~al.(2020)Rasley, Rajbhandari, Ruwase, and
  He}]{DBLP:conf/kdd/RasleyRRH20}
Jeff Rasley, Samyam Rajbhandari, Olatunji Ruwase, and Yuxiong He. 2020.
\newblock Deepspeed: System optimizations enable training deep learning models
  with over 100 billion parameters.
\newblock In \emph{{KDD} '20: The 26th {ACM} {SIGKDD} Conference on Knowledge
  Discovery and Data Mining, Virtual Event, CA, USA, August 23-27, 2020}, pages
  3505--3506. {ACM}.

\bibitem[{Razniewski et~al.(2016)Razniewski, Suchanek, and
  Nutt}]{DBLP:conf/akbc/RazniewskiSN16}
Simon Razniewski, Fabian~M. Suchanek, and Werner Nutt. 2016.
\newblock But what do we actually know?
\newblock In \emph{Proceedings of the 5th Workshop on Automated Knowledge Base
  Construction, AKBC@NAACL-HLT 2016, San Diego, CA, USA, June 17, 2016}, pages
  40--44. The Association for Computer Linguistics.

\bibitem[{Reddy et~al.(2019)Reddy, Chen, and
  Manning}]{DBLP:journals/tacl/ReddyCM19}
Siva Reddy, Danqi Chen, and Christopher~D. Manning. 2019.
\newblock Coqa: {A} conversational question answering challenge.
\newblock \emph{Trans. Assoc. Comput. Linguistics}, 7:249--266.

\bibitem[{Scao et~al.(2022)Scao, Fan, Akiki, Pavlick, Ilic, Hesslow,
  Castagn{\'{e}}, Luccioni, Yvon, Gall{\'{e}}, Tow, Rush, Biderman, Webson,
  Ammanamanchi, Wang, Sagot, Muennighoff, del Moral, Ruwase, Bawden, Bekman,
  McMillan{-}Major, Beltagy, Nguyen, Saulnier, Tan, Suarez, Sanh,
  Lauren{\c{c}}on, Jernite, Launay, Mitchell, Raffel, Gokaslan, Simhi, Soroa,
  Aji, Alfassy, Rogers, Nitzav, Xu, Mou, Emezue, Klamm, Leong, van Strien,
  Adelani, and et~al.}]{DBLP:journals/corr/abs-2211-05100}
Teven~Le Scao, Angela Fan, Christopher Akiki, Ellie Pavlick, Suzana Ilic,
  Daniel Hesslow, Roman Castagn{\'{e}}, Alexandra~Sasha Luccioni,
  Fran{\c{c}}ois Yvon, Matthias Gall{\'{e}}, Jonathan Tow, Alexander~M. Rush,
  Stella Biderman, Albert Webson, Pawan~Sasanka Ammanamanchi, Thomas Wang,
  Beno{\^{\i}}t Sagot, Niklas Muennighoff, Albert~Villanova del Moral, Olatunji
  Ruwase, Rachel Bawden, Stas Bekman, Angelina McMillan{-}Major, Iz~Beltagy,
  Huu Nguyen, Lucile Saulnier, Samson Tan, Pedro~Ortiz Suarez, Victor Sanh,
  Hugo Lauren{\c{c}}on, Yacine Jernite, Julien Launay, Margaret Mitchell, Colin
  Raffel, Aaron Gokaslan, Adi Simhi, Aitor Soroa, Alham~Fikri Aji, Amit
  Alfassy, Anna Rogers, Ariel~Kreisberg Nitzav, Canwen Xu, Chenghao Mou, Chris
  Emezue, Christopher Klamm, Colin Leong, Daniel van Strien, David~Ifeoluwa
  Adelani, and et~al. 2022.
\newblock {BLOOM:} {A} 176b-parameter open-access multilingual language model.
\newblock \emph{CoRR}, abs/2211.05100.

\bibitem[{Sevegnani et~al.(2021)Sevegnani, Howcroft, Konstas, and
  Rieser}]{DBLP:conf/acl/SevegnaniHKR20}
Karin Sevegnani, David~M. Howcroft, Ioannis Konstas, and Verena Rieser. 2021.
\newblock Otters: One-turn topic transitions for open-domain dialogue.
\newblock In \emph{Proceedings of the 59th Annual Meeting of the Association
  for Computational Linguistics and the 11th International Joint Conference on
  Natural Language Processing, {ACL/IJCNLP} 2021, (Volume 1: Long Papers),
  Virtual Event, August 1-6, 2021}, pages 2492--2504. Association for
  Computational Linguistics.

\bibitem[{Shuster et~al.(2022)Shuster, Komeili, Adolphs, Roller, Szlam, and
  Weston}]{DBLP:journals/corr/abs-2203-13224}
Kurt Shuster, Mojtaba Komeili, Leonard Adolphs, Stephen Roller, Arthur Szlam,
  and Jason Weston. 2022.
\newblock Language models that seek for knowledge: Modular search {\&}
  generation for dialogue and prompt completion.
\newblock \emph{CoRR}, abs/2203.13224.

\bibitem[{Stede and Schlangen(2004)}]{stede2004information}
Manfred Stede and David Schlangen. 2004.
\newblock Information-seeking chat: Dialogues driven by topic-structure.
\newblock In \emph{Proceedings of Catalog (the 8th workshop on the semantics
  and pragmatics of dialogue; SemDial04)}. Citeseer.

\bibitem[{Swart et~al.(2017)Swart, Peters, and
  Broersma}]{swart2017repositioning}
Joelle Swart, Chris Peters, and Marcel Broersma. 2017.
\newblock Repositioning news and public connection in everyday life: A
  user-oriented perspective on inclusiveness, engagement, relevance, and
  constructiveness.
\newblock \emph{Media, culture \& society}, 39(6):902--918.

\bibitem[{Tang et~al.(2020)Tang, Tran, Li, Chen, Goyal, Chaudhary, Gu, and
  Fan}]{DBLP:journals/corr/abs-2008-00401}
Yuqing Tang, Chau Tran, Xian Li, Peng{-}Jen Chen, Naman Goyal, Vishrav
  Chaudhary, Jiatao Gu, and Angela Fan. 2020.
\newblock Multilingual translation with extensible multilingual pretraining and
  finetuning.
\newblock \emph{CoRR}, abs/2008.00401.

\bibitem[{Thoppilan et~al.(2022)Thoppilan, De~Freitas, Hall, Shazeer,
  Kulshreshtha, Cheng, Jin, Bos, Baker, Du et~al.}]{thoppilan2022lamda}
Romal Thoppilan, Daniel De~Freitas, Jamie Hall, Noam Shazeer, Apoorv
  Kulshreshtha, Heng-Tze Cheng, Alicia Jin, Taylor Bos, Leslie Baker, Yu~Du,
  et~al. 2022.
\newblock Lamda: Language models for dialog applications.
\newblock \emph{arXiv preprint arXiv:2201.08239}.

\bibitem[{Wang et~al.(2022)Wang, Zhang, Zhang, Yang, Gao, Wu, Dong, He, Zhuo,
  Yang, Huang, Li, Wu, Lu, Zhu, Chen, Han, Pan, Wang, Wang, Wu, Zeng, Chen,
  Gan, and Zhang}]{fengshenbang}
Junjie Wang, Yuxiang Zhang, Lin Zhang, Ping Yang, Xinyu Gao, Ziwei Wu, Xiaoqun
  Dong, Junqing He, Jianheng Zhuo, Qi~Yang, Yongfeng Huang, Xiayu Li, Yanghan
  Wu, Junyu Lu, Xinyu Zhu, Weifeng Chen, Ting Han, Kunhao Pan, Rui Wang, Hao
  Wang, Xiaojun Wu, Zhongshen Zeng, Chongpei Chen, Ruyi Gan, and Jiaxing Zhang.
  2022.
\newblock Fengshenbang 1.0: Being the foundation of chinese cognitive
  intelligence.
\newblock \emph{CoRR}, abs/2209.02970.

\bibitem[{Wang et~al.(2020)Wang, Ke, Zheng, Huang, Jiang, Zhu, and
  Huang}]{wang2020chinese}
Yida Wang, Pei Ke, Yinhe Zheng, Kaili Huang, Yong Jiang, Xiaoyan Zhu, and
  Minlie Huang. 2020.
\newblock A large-scale chinese short-text conversation dataset.
\newblock In \emph{NLPCC}.

\bibitem[{Wolf et~al.(2020)Wolf, Debut, Sanh, Chaumond, Delangue, Moi, Cistac,
  Rault, Louf, Funtowicz, Davison, Shleifer, von Platen, Ma, Jernite, Plu, Xu,
  Scao, Gugger, Drame, Lhoest, and Rush}]{DBLP:conf/emnlp/WolfDSCDMCRLFDS20}
Thomas Wolf, Lysandre Debut, Victor Sanh, Julien Chaumond, Clement Delangue,
  Anthony Moi, Pierric Cistac, Tim Rault, R{\'{e}}mi Louf, Morgan Funtowicz,
  Joe Davison, Sam Shleifer, Patrick von Platen, Clara Ma, Yacine Jernite,
  Julien Plu, Canwen Xu, Teven~Le Scao, Sylvain Gugger, Mariama Drame, Quentin
  Lhoest, and Alexander~M. Rush. 2020.
\newblock Transformers: State-of-the-art natural language processing.
\newblock In \emph{Proceedings of the 2020 Conference on Empirical Methods in
  Natural Language Processing: System Demonstrations, {EMNLP} 2020 - Demos,
  Online, November 16-20, 2020}, pages 38--45. Association for Computational
  Linguistics.

\bibitem[{Wu et~al.(2019)Wu, Guo, Zhou, Wu, Zhang, Lian, and
  Wang}]{DBLP:conf/acl/WuGZWZLW19}
Wenquan Wu, Zhen Guo, Xiangyang Zhou, Hua Wu, Xiyuan Zhang, Rongzhong Lian, and
  Haifeng Wang. 2019.
\newblock Proactive human-machine conversation with explicit conversation goal.
\newblock In \emph{Proceedings of the 57th Conference of the Association for
  Computational Linguistics, {ACL} 2019, Florence, Italy, July 28- August 2,
  2019, Volume 1: Long Papers}, pages 3794--3804. Association for Computational
  Linguistics.

\bibitem[{Wu et~al.(2022)Wu, Parish, Cheng, Min, Ammanabrolu, Ostendorf, and
  Hajishirzi}]{DBLP:journals/corr/abs-2207-00746}
Zeqiu Wu, Ryu Parish, Hao Cheng, Sewon Min, Prithviraj Ammanabrolu, Mari
  Ostendorf, and Hannaneh Hajishirzi. 2022.
\newblock {INSCIT:} information-seeking conversations with mixed-initiative
  interactions.
\newblock \emph{CoRR}, abs/2207.00746.

\bibitem[{Xue et~al.(2020)Xue, Constant, Roberts, Kale, Al-Rfou, Siddhant,
  Barua, and Raffel}]{xue2020mt5}
Linting Xue, Noah Constant, Adam Roberts, Mihir Kale, Rami Al-Rfou, Aditya
  Siddhant, Aditya Barua, and Colin Raffel. 2020.
\newblock mt5: A massively multilingual pre-trained text-to-text transformer.
\newblock \emph{arXiv preprint arXiv:2010.11934}.

\bibitem[{Zhang et~al.(2022)Zhang, Roller, Goyal, Artetxe, Chen, Chen, Dewan,
  Diab, Li, Lin, Mihaylov, Ott, Shleifer, Shuster, Simig, Koura, Sridhar, Wang,
  and Zettlemoyer}]{DBLP:journals/corr/abs-2205-01068}
Susan Zhang, Stephen Roller, Naman Goyal, Mikel Artetxe, Moya Chen, Shuohui
  Chen, Christopher Dewan, Mona Diab, Xian Li, Xi~Victoria Lin, Todor Mihaylov,
  Myle Ott, Sam Shleifer, Kurt Shuster, Daniel Simig, Punit~Singh Koura, Anjali
  Sridhar, Tianlu Wang, and Luke Zettlemoyer. 2022.
\newblock {OPT:} open pre-trained transformer language models.
\newblock \emph{CoRR}, abs/2205.01068.

\bibitem[{Zhang et~al.(2019)Zhang, Feng, Meng, You, and
  Liu}]{DBLP:conf/acl/ZhangFMYL19}
Wen Zhang, Yang Feng, Fandong Meng, Di~You, and Qun Liu. 2019.
\newblock Bridging the gap between training and inference for neural machine
  translation.
\newblock In \emph{Proceedings of the 57th Conference of the Association for
  Computational Linguistics, {ACL} 2019, Florence, Italy, July 28- August 2,
  2019, Volume 1: Long Papers}, pages 4334--4343. Association for Computational
  Linguistics.

\bibitem[{Zhang et~al.(2021)Zhang, Zhang, Chen, Guo, Hua, Wang, and
  Zhou}]{zhang2021mengzi}
Zhuosheng Zhang, Hanqing Zhang, Keming Chen, Yuhang Guo, Jingyun Hua, Yulong
  Wang, and Ming Zhou. 2021.
\newblock \href {http://arxiv.org/abs/2110.06696} {Mengzi: Towards lightweight
  yet ingenious pre-trained models for chinese}.

\bibitem[{Zhao et~al.(2019)Zhao, Chen, Zhang, Zhao, Liu, Lu, Chen, Deng, Ju,
  and Du}]{DBLP:conf/emnlp/ZhaoCZZLLCDJD19}
Zhe Zhao, Hui Chen, Jinbin Zhang, Xin Zhao, Tao Liu, Wei Lu, Xi~Chen, Haotang
  Deng, Qi~Ju, and Xiaoyong Du. 2019.
\newblock {UER:} an open-source toolkit for pre-training models.
\newblock In \emph{Proceedings of the 2019 Conference on Empirical Methods in
  Natural Language Processing and the 9th International Joint Conference on
  Natural Language Processing, {EMNLP-IJCNLP} 2019, Hong Kong, China, November
  3-7, 2019 - System Demonstrations}, pages 241--246. Association for
  Computational Linguistics.

\bibitem[{Zhou et~al.(2020)Zhou, Zheng, Huang, Huang, and
  Zhu}]{DBLP:conf/acl/ZhouZHHZ20}
Hao Zhou, Chujie Zheng, Kaili Huang, Minlie Huang, and Xiaoyan Zhu. 2020.
\newblock Kdconv: {A} chinese multi-domain dialogue dataset towards multi-turn
  knowledge-driven conversation.
\newblock In \emph{Proceedings of the 58th Annual Meeting of the Association
  for Computational Linguistics, {ACL} 2020, Online, July 5-10, 2020}, pages
  7098--7108. Association for Computational Linguistics.

\bibitem[{Zhou et~al.(2018)Zhou, Prabhumoye, and
  Black}]{DBLP:conf/emnlp/ZhouPB18}
Kangyan Zhou, Shrimai Prabhumoye, and Alan~W. Black. 2018.
\newblock A dataset for document grounded conversations.
\newblock In \emph{Proceedings of the 2018 Conference on Empirical Methods in
  Natural Language Processing, Brussels, Belgium, October 31 - November 4,
  2018}, pages 708--713. Association for Computational Linguistics.

\bibitem[{Zhou et~al.(2022)Zhou, Gopalakrishnan, Hedayatnia, Kim, Pujara, Ren,
  Liu, and Hakkani{-}Tur}]{DBLP:conf/acl/Zhou0HKP0LH22}
Pei Zhou, Karthik Gopalakrishnan, Behnam Hedayatnia, Seokhwan Kim, Jay Pujara,
  Xiang Ren, Yang Liu, and Dilek Hakkani{-}Tur. 2022.
\newblock Think before you speak: Explicitly generating implicit commonsense
  knowledge for response generation.
\newblock In \emph{Proceedings of the 60th Annual Meeting of the Association
  for Computational Linguistics (Volume 1: Long Papers), {ACL} 2022, Dublin,
  Ireland, May 22-27, 2022}, pages 1237--1252. Association for Computational
  Linguistics.

\end{thebibliography}
\bibliographystyle{acl_natbib}

\appendix

\clearpage

\label{sec:appendix}

\section{Case Study}
\begin{figure*}
    \centering
    \includegraphics[width=\textwidth]{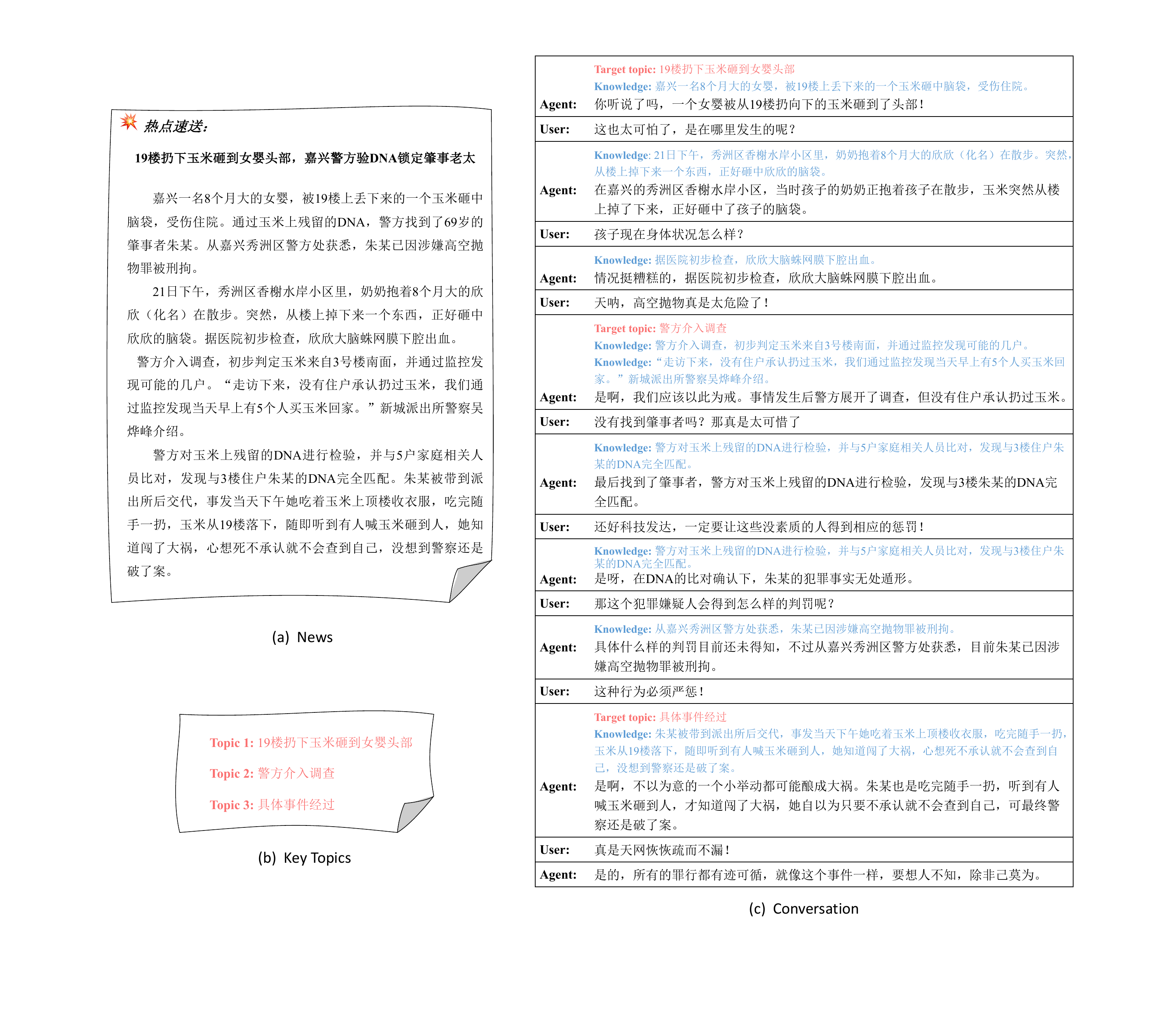}
    \caption{An example of \ts{NewsDialogues}. For reading conveniently, we translate the original Chinese dialogue to English and omit some information in Figure \ref{fig:example}. Here is the original version. During the long conversation, the agent proactively steers the conversation to the key topics of news.}
    \label{fig:chinese_example}
    \vsd
\end{figure*}

For reading convenience, we translate the original Chinese conversation to its English version in Figure \ref{fig:example}.
Take the original version in Figure \ref{fig:chinese_example}.

\section{Annotator Profile}
We employ 30 crowdworkers with equally distributed genders for our annotations.
They are all native Chinese speakers with ages from 20 to 40 years old.
In addition, they are from different regions of China.
We pay them a wage above the average in their area.
It takes 180,000 ChineseYuan (CNY) for constructing \ts{NewsDialogues}.

\section{Implementation Details}
\label{app:implementation_details}
All our experiments are based on Transformers\footnote{\url{https://huggingface.co/docs/transformers/index}} \cite{DBLP:conf/emnlp/WolfDSCDMCRLFDS20}, DeepSpeed\footnote{\url{https://github.com/microsoft/DeepSpeed}} \cite{DBLP:conf/kdd/RasleyRRH20} and Pytorch Lightning\footnote{\url{https://github.com/Lightning-AI/lightning}}.

\ntb{General Setting.}
\begin{table}[t]
    \begin{center}
    \centering
    \small
    \resizebox{\linewidth}{!}{%
    \begin{tabular}{lcc}
    \toprule
       \tb{Model} & \tb{MSL} & \tb{\#Params } \\
    \midrule
    \midrule
        CDial-GPT \citep{wang2020chinese} & 512 & 95.5M \\
        EVA2.0 \citep{DBLP:journals/corr/abs-2203-09313} & 512 & 970M \\
        BLOOM \citep{DBLP:journals/corr/abs-2211-05100} & 2048 & 560M \\
        mBART \citep{DBLP:journals/corr/abs-2008-00401} & 1024 & 610M \\
        mT5 \citep{xue2020mt5} & 2048 & 582M \\
        Chinese GPT \citep{DBLP:conf/emnlp/ZhaoCZZLLCDJD19} & 1024 & 102M \\
        Chinese BART \citep{fengshenbang} & 1024 & 759M \\
        Chinese T5 \citep{fengshenbang} & 2048 & 784M \\
        Predict-Generate-Rank (Ours) & 2048 & 784M + 102M \\
    \bottomrule
    \end{tabular}
    }
    \end{center}

    \caption{
        The maximum sequence length (MSL) and the parameter number of each model.
    }
    \label{tab:model}
    \vspace{-5pt}
\end{table}
For both encoder-decoder models and decoder-only models, the input sequence is the concatenation of the news, key topics and the dialogue history, the output sequence is the concatenation of the grounded knowledge and response.
We truncate the input sequence according to the maximum sequence length of the model when it uses absolute position embedding. 
For the T5-based models with relative position embedding, we set the maximum sequence length as 2048.
The maximum sequence length and parameters of each model are shown in Table \ref{tab:model}.
All generative models follow the same hyper-parameter setting.
For training, we set the learning rate as $5e-5$, batch size as $32$, and use Adam optimizer \citep{DBLP:journals/corr/KingmaB14} with warmup learning rate schedule, the warmup ratio is $0.1$.
Each model is trained for 2k gradient steps, and we choose the checkpoint with the lowest perplexity score on the validation set for evaluation.
For generation, we use Top-\ti{p} sampling \cite{DBLP:conf/iclr/HoltzmanBDFC20} with p=0.9.
We run all experiments three times and report the best results in this paper.

\ntb{Ours.}
Our generator is trained with the same hyper-parameter setting as above.
For the ranker, the learning rate and batch size are 5e-5 and 64 respectively.
The optimizer is the same as that of the generator.
We set the maximum gradient steps as 20K for the pretraining stage and 10K for the finetuning stage, the checkpoint with the highest accuracy on the validation set is used for evaluation.
After processing DuConv and KdConv, we have $257146$ examples for pre-training the ranker, where each example has 1 positive response and 7 negative responses which are randomly sampled from the datasets.
We randomly split these examples with a ratio of $4: 1$ for the training and validation processes of the pre-training stage.
For finetuning the ranker on \ts{NewsDialogues}, we predict 96 grounded knowledge for each example in the training set of \ts{NewsDialogues} and generate responses based on them, finally we can get $597504$ responses. 
Then, we construct positive responses and negative responses based on $\gamma_1=50$ and $\gamma_2=15$, each positive response is paired with 7 negative responses as in the pre-trainig stage.
Totally, We can get $35159$ examples for the training process of the finetuning stage.
Using the same method on the validation set of \ts{NewsDialogues}, we can get 2854 examples for the validation process of the finetuning stage.
Our ranker gets 91.73 accuracy at the pre-training stage and 59.28 accuracy at the finetuning stage.

\section{Human Interactive Evaluation Setting}
\label{app:human_interactive_evaluation_setting}
We employ 4 humans for human interactive evaluation and collect $40$ conversations for each model.
Specifically, each conversation is grounded on a news article from the test set of \ts{NewsDialogues}, and contains at least 10 turns, 5 from the human and 5 from the model.
In addition, we also select the $40$ conversations with the same news articles from the test set to further investigate the performance gap between humans and current models.
In total, we have $120$ conversations, which are then distributed to 4
human evaluators to score from various aspects.

\end{document}